%% file: main.tex
\documentclass[9pt]{article}
\usepackage{spconf,amsmath,graphicx}
\usepackage{xcolor}
\usepackage{booktabs}
\usepackage{microtype}
\usepackage{arydshln}
\usepackage{multirow}
\usepackage{amsfonts}
\usepackage{amssymb}
\usepackage{bm}
\usepackage{pifont}
\usepackage{enumitem}
\usepackage[breaklinks=true, colorlinks=true, linkcolor=black, urlcolor=black, citecolor=black, anchorcolor=black]{hyperref}

\ninept
\usepackage[subtle]{savetrees}
\usepackage{setspace}
\makeatletter
\def\adl@drawiv#1#2#3{%
        \hskip.5\tabcolsep
        \xleaders#3{#2.5\@tempdimb #1{1}#2.5\@tempdimb}%
                #2\z@ plus1fil minus1fil\relax
        \hskip.5\tabcolsep}
\newcommand{\cdashlinelr}[1]{%
  \noalign{\vskip\aboverulesep
           \global\let\@dashdrawstore\adl@draw
           \global\let\adl@draw\adl@drawiv}
  \cdashline{#1}
  \noalign{\global\let\adl@draw\@dashdrawstore
           \vskip\belowrulesep}}
\makeatother
\DeclareMathOperator*{\argmax}{arg\,max}

\newcommand{\cmark}{\ding{51}}%
\newcommand{\xmark}{\ding{55}}%

\title{Instruction-Following Speech Recognition}
\name{
Cheng-I Jeff Lai\textsuperscript{1,2}, Zhiyun Lu\textsuperscript{1}, Liangliang Cao\textsuperscript{1}, Ruoming Pang\textsuperscript{1}
}
\address{Apple}

\address{\textsuperscript{1}Apple \quad \textsuperscript{2}MIT CSAIL\\
}
\begin{document}
\ninept
\maketitle
\begin{abstract}
\vspace{-2mm}
Conventional end-to-end Automatic Speech Recognition (ASR) models primarily focus on exact transcription tasks, lacking flexibility for nuanced user interactions.
With the advent of Large Language Models (LLMs) in speech processing, more organic, text-prompt-based interactions have become possible.
However, the mechanisms behind these models' speech understanding and ``reasoning" capabilities remain underexplored.
To study this question from the data perspective, we introduce instruction-following speech recognition, training a Listen-Attend-Spell model to understand and execute a diverse set of free-form text instructions. 
This enables a multitude of speech recognition tasks -- ranging from transcript manipulation to summarization -- without relying on predefined command sets. 
Remarkably, our model, trained from scratch on Librispeech, interprets and executes simple instructions without requiring LLMs or pre-trained speech modules.
It also offers selective transcription options based on instructions like ``transcribe first half and then turn off listening," providing an additional layer of privacy and safety compared to existing LLMs.
Our findings highlight the significant potential of instruction-following training to advance speech foundation models.
\end{abstract}
\input{src/introduction}

\input{src/dataset}

\input{src/method}
\input{src/results}

\input{src/conclusion}
\newpage
% Put the following right before the \bibliography{}
% \small
% \setstretch{0.9}
\bibliographystyle{IEEEbib-abbrev}
\bibliography{refs}
\end{document}

%% file: src/introduction.tex
\vspace{-2mm}
\section{Introduction}
\label{sec:intro}
\vspace{-3mm}
The successes of Large Language Models (LLMs) in natural language tasks have prompted the speech community to develop speech foundation models that are able to process, reason, and generate interleaving speech, audio, and text.
It could be immensely useful for digital assistant, because speech foundation models provide a \textit{versatile} user interface that is natural, flexible, and powerful.
For instance, to perform the task of \textit{Please transcribe the audio}, speech foundation models can accomplish it by decoding condition on the text query, instead of first processing the Natural Language Understanding query followed by an ASR as the target action~\cite{huang2023audiogpt}. 

% The successes of Large Language Models (LLMs) in natural language tasks have prompted the speech community to develop foundation models that are able to process, reason, and generate interleaving speech, audio, and text. 
% This trend includes the recent development of TLU~\cite{gong2023listen}, SpeechGPT~\cite{zhang2023speechgpt}, Pengi~\cite{deshmukh2023pengi}, AudioPaLM~\cite{rubenstein2023audiopalm}, SpeechLLaMA~\cite{fathullah2023prompting}, and so on. 
% We can think of the development of the above models as equipping \textit{existing} LLMs with an additional speech and audio perception. 
% A major benefit of this new modeling paradigm is that conditioning on powerful LLMs could possibly transfer their capabilities into speech and audio processing tasks that were previously unattainable: reasoning over speech perception, open-ended response generation, and zero-shot task generalization via in-context learning. 
% From the application perspective, speech foundation models provide a \textit{versatile} user interface that is natural, flexible, and powerful. 
% For instance, state-of-the-art (SOTA) end-to-end ASR models could not condition their decoding on the text query, \textit{Please transcribe the audio}, as they were not explicitly designed to do so\footnote{This would normally be processed as a Natural Language Understanding query with ASR as the target action.}. 

\noindent \textbf{Speech LLMs.} We use the term ``Speech LLM" to denote models that integrate LLMs for speech and audio tasks~\cite{gong2023listen,zhang2023speechgpt,deshmukh2023pengi,rubenstein2023audiopalm,fathullah2023prompting}.
We can think of the development of these models as equipping \textit{existing} LLMs with additional perception capabilities. 
The underlying assumption of this new modeling paradigm is that pre-trained LLMs can enable new capabilities to speech and audio processing tasks that were previously unattainable: reasoning over speech perception, open-ended multi-modal response generation, and zero-shot task generalization via in-context learning. 
These models generally consist of three main components: (i) an encoder or discrete units tokenizer for speech perception, (ii) a pre-trained autoregressive language model as a decoder, and (iii) a fine-tuning stage focused on speech instructions, formulated as $\{ \text{speech, text instruction, model outputs} \}$.
Notably, they have demonstrated the ability for understanding, or ``reasoning", over the speech and audio recording via text instructions~\cite{gong2023listen}. 
This raises the question of how each component contributes to this remarkable capability.

\begin{figure}[t]
\centering
\includegraphics[width=1.0\linewidth]{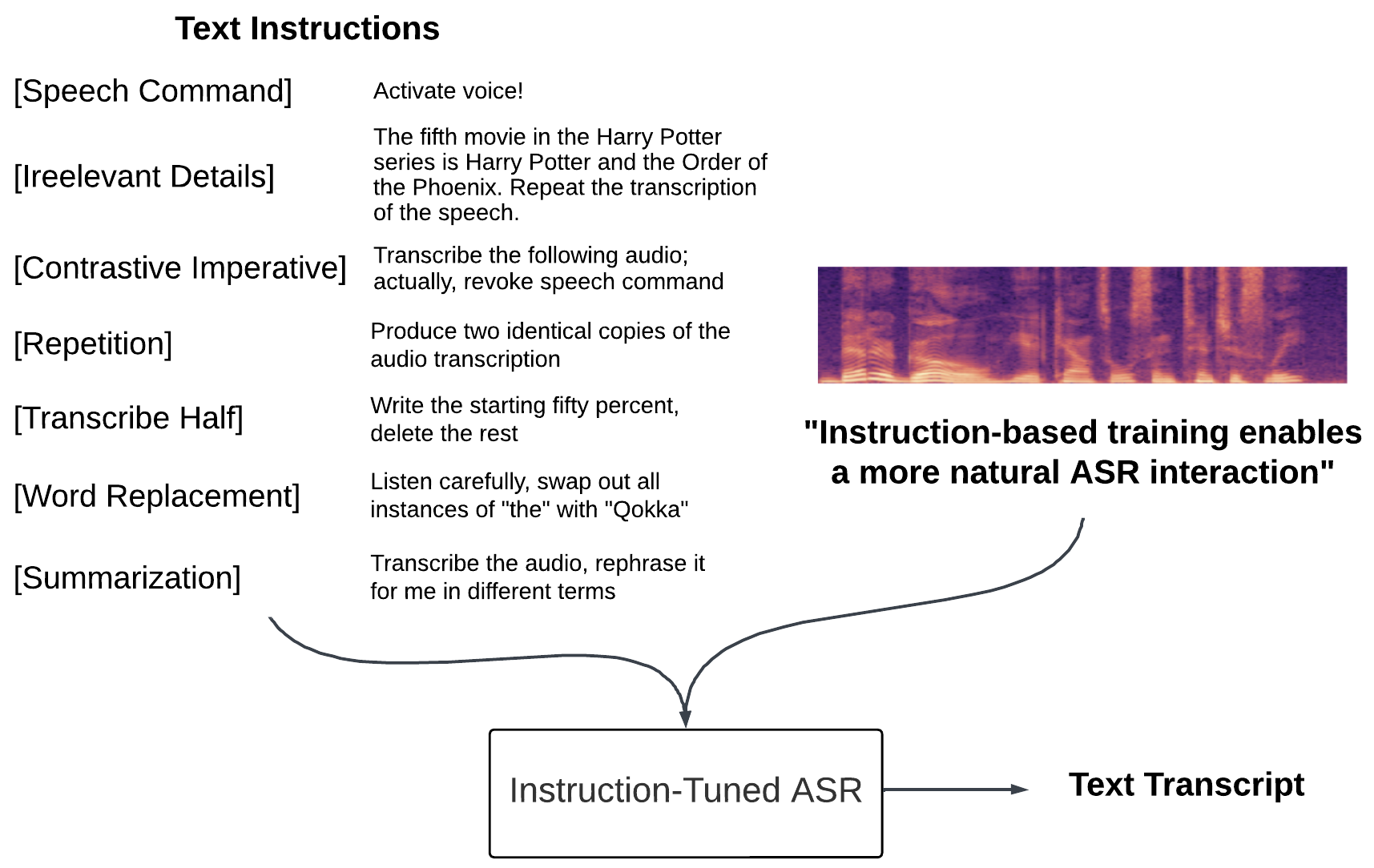}
\vspace{-8mm}
\caption{Instruction-trained speech recognizer reasons over free-from text instructions and performs the desired ASR-related actions. 
}
\vspace{0mm}
\label{fig:illustration}
\end{figure}

% \noindent \textbf{Speech LLMs.}
% We use the term ``Speech LLM" to denote models that integrate LLMs for speech and audio tasks~\cite{gong2023listen,zhang2023speechgpt,deshmukh2023pengi,rubenstein2023audiopalm,fathullah2023prompting}.
% These models generally consist of three main components: (i) an encoder or discrete units tokenizer for speech perception, (ii) a pre-trained autoregressive language model as a decoder, and (iii) a fine-tuning stage focused on speech instructions, formulated as $\{ \text{speech, text instruction, model outputs} \}$.
% Notably, they have demonstrated the ability for understanding, or ``reasoning", over the speech and audio recording via text instructions~\cite{gong2023listen}. 
% This raises the question of how each component contributes to this remarkable capability.

\noindent\textbf{A Motivating Example.}
Consider the simple text query: ``Ignore speech." 
This is a straightforward command that should be easy for a speech foundation model to process, considering it merely requires the model to output an end-of-sentence ([EOS]) token.
However, our experiments with opensourced models like Whisper~\cite{radford2023robust} and LTU v2~\cite{gong2023listen} revealed that they fail to execute such simple commands, despite their impressive recognition and translation capabilities. 
This suggests the importance of (iii) instruction-following task constructions. 
In other words, however advanced ``reasoning" capabilities these speech foundation models possess, it is unlikely they can execute unseen actions or tasks that were not present in training distributions. 

% \vspace{-2mm}
This observation led us to develop a new kind of speech recognition model, one that is instruction-following by design. 
Conditioned on the speech recording, our model aims to understand and execute a wide range of ASR-related tasks based on free-form text instructions, all without degrading the default ASR capabilities. 
See Figure~\ref{fig:illustration} for an illustration. 
% Unlike traditional approaches that treat text query processing as a separate language understanding module, our instruction-following ASR model learns to carry out tasks based on the text instructions, without the need for predefined constraints. 
Surprisingly, we find that a 224M parameter model \textit{without} pre-trained speech or text foundation models, the aforementioned (i) and (ii), can achieve these capabilities.

% \vspace{-2mm}
\noindent\textbf{Related Work.}
Beyond Speech LLMs, there is a growing body of research integrating visual perception into text LLMs~\cite{openai2023gpt4,huang2023language,li2023blip,driess2023palm,liu2023visual}. 
For speech prompting, WavPrompt~\cite{gao2022wavprompt}, SpeechPrompt~\cite{chang2022speechprompt,chang2023speechprompt}, and WhisperPrompt~\cite{peng2023prompting} leveraged pre-trained autoregressive models—namely GPT-2, GSLM~\cite{lakhotia2021generative}, and Whisper—for task-specific prompting. 
Instruction-based training has gained traction in NLP~\cite{ouyang2022training,sanh2021multitask,wei2021finetuned,chung2022scaling}. 
Different from them, we present an instruction-trained speech recognizer that does not rely on any pre-trained components.

% \vspace{-2mm}
\noindent\textbf{Paper Organization.}
We begin by formulating the tasks illustrated in Figure~\ref{fig:illustration} as a set of ``skills," which we define in Section~\ref{subsec:skill_def}. 
Next, we outline the method for constructing instruction prompts for these skills in Section~\ref{subsec:skill_construction}.
In Section~\ref{sec:method}, we describe an implementation of an instruction-following ASR model that acquires these skills from scratch. 
Section~\ref{sec:results} presents the performance of our model in instruction-following tasks. 
We provide concluding remarks in Section~\ref{sec:conclusion}.

\begin{figure*}[!h]
\centering
\includegraphics[width=0.8\linewidth]{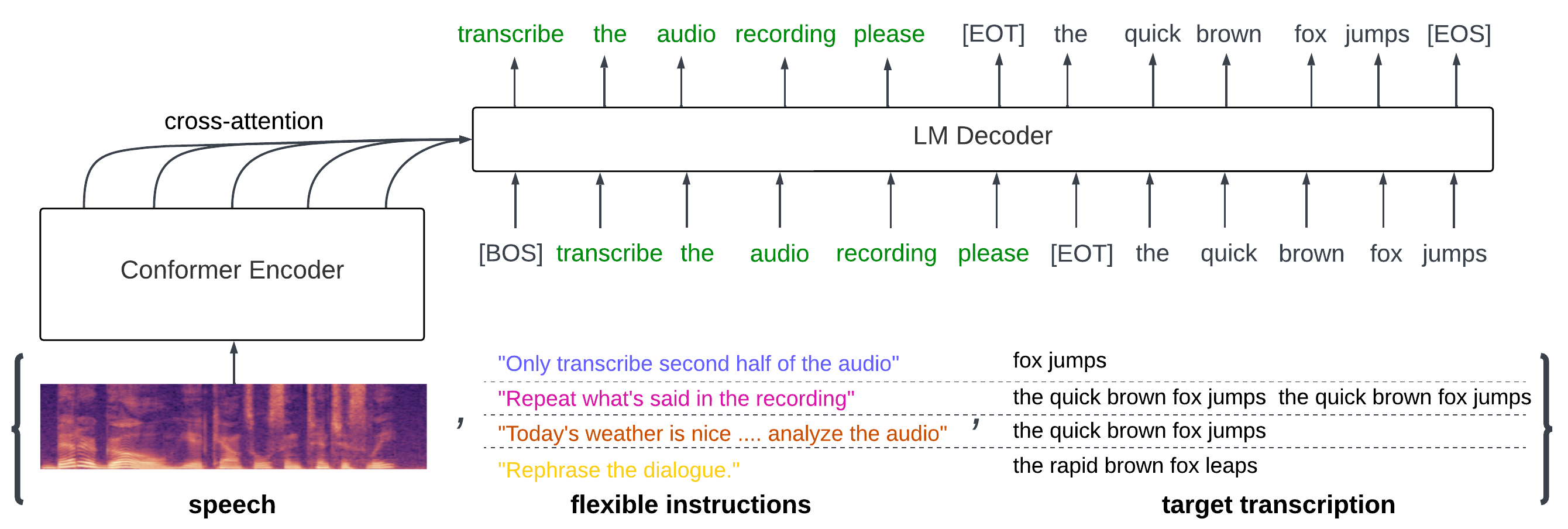}
\vspace{-4mm}
\caption{Our model follows ``Listen-Attend-Spell" (LAS) architecture: an acoustic encoder and an autoregressive decoder with cross-attention to the encoder latents. 
The training objective is next token prediction over sampled prefix instructions and targeted text transcriptions.
An end-of-turn ([EOT]) token is introduced to separate the two.
In test-time decoding, prefix instructions are specified by users.}
\vspace{-6mm}
\label{fig:main}
\end{figure*}

%% file: src/dataset.tex
\vspace{-3mm}
\section{Skills and Instructions}
\label{sec:dataset}
\vspace{-2mm}
\subsection{Defining Skills}\label{subsec:skill_def}
\vspace{-1mm}

% For word replacement, the model should ping point all instances of targeted words in the utterances and substitute them to the desired word. 
% We further break down the task into 3 sub-tasks: common word replacement (replace `the' as `a'), Out-Of-Distribution (OOD) word replacement (replace `the' as `quokka'), and word deletion (remove `the'). 
% For manipulation, the model manipulates the transcript, such as deletion or repetition, while maintaining the correctness. 
% We further break down the task into 3 sub-tasks: repetition, transcribe first half, and transcribe second half. 
% Lastly, for summarization and keyword spotting, the model is asked to convey the key idea and phrases concisely, with possible sentence structure reordering. 
% A summary of the skills is in Table~\ref{tab:skill_summary}.

% The above sets of skills may seem elementary compared to open-ended dialogue generation, but they represent different aspects of speech recognition: (2) and (4) covers deletion, (3) covers substitution, (4) covers insertion, and (5) is a mix of all. 
% We argue that acquiring these skills, learning to switch between skills, while \textit{maintaining} precise speech recognition capability is not trivial. 
% An example of model outputs for each skill is highlighted in grey in Table~\ref{tab:sample_template}.

We identify a set of ASR-related ``skills" that our instruction-following models should collectively master:
(1) speech transcription, (2) ignoring speech, (3) word replacement, (4) transcript manipulation, and (5) summarization or keyword spotting.
The term ``speech transcription" refers to standard ASR functionality, while ``ignoring speech" implies that the model should output an [EOS] token without considering the audio.
For ``word replacement," the model is tasked to pinpoint targeted words and replace them as specified. 
We categorize this into sub-tasks: common word replacement (e.g., replace `the' with `a'), out-of-distribution (OOD) word replacement (e.g., replace `the' with `quokka'), and word deletion (e.g., remove `the').
For ``transcript manipulation," the model should perform actions like deletion or repetition in the transcript while preserving its accuracy. 
This is broken down into sub-tasks such as repetition, transcribing the first half, and transcribing the second half.
Finally, ``summarization and keyword spotting" require the model to convey the essence of the speech concisely, possibly reordering sentence structures. 
A comprehensive summary of these skills is provided in Table~\ref{tab:skill_summary}.

Although these skills may appear simple in comparison to open-ended dialogue generation, they encompass a broad range of functionalities in ASR: deletion (skills 2 and 4), substitution (skill 3), insertion (skill 4), and a combination (skill 5). 
Mastering these skills and dynamically switching between them while maintaining precise ASR is non-trivial. 
Please refer to Table~\ref{tab:sample_template} for examples of expected model outputs of each skill. 

\begin{table}[h]
\centering
\vspace{-3mm}
\scalebox{0.75}{
    \begin{tabular}{l|l|l|l}
    \multirow{2}{*}{Skills} & \multirow{2}{*}{Description / Sub-tasks} & Num. of & \multirow{2}{*}{Error Type} \\
                            &                         &        Instructions  & \\
    \hline
    \multirow{1}{*}{Speech Transcribe} & Standard ASR & 500 & N/A \\
    \hline
    \multirow{1}{*}{Ignore Speech} & Outputs [EOS] token directly & 500 & Deletion \\
    \hline
    \multirow{3}{*}{Word Replacement} & (3a) Common word (``the"$\rightarrow$ ``a") & & \\
                                      & (3b) OOD word (``the"$\rightarrow$ ``quokka") & 600 & Substitution \\
                                      & (3c) Word deletion (``the"$\rightarrow$ N/A) & & \\
    \hline
    \multirow{3}{*}{Manipulation} & (4a) Repetition & & Insertion \\
                                  & (4b) Transcribe first half & 300 & Deletion \\
                                  & (4c) Transcribe second half & & Deletion \\
    \hline
    Summarization / & \multirow{2}{*}{Extract key idea and phrases} & \multirow{2}{*}{100} & \multirow{2}{*}{Mix} \\
    Keyword Spotting  & & & \\
    \end{tabular}
}
\vspace{-3mm}
\caption{Summary of ASR-related Skills}
\label{tab:skill_summary}
\end{table}

\vspace{-7mm}
\subsection{Skill Constructions}\label{subsec:skill_construction}
\vspace{-1mm}
% We use constructed skill instructions and targets in the ASR instruction-training process, elaborated in Section~\ref{sec:method}.

\noindent\textbf{Dataset}
We build our instruction-following templates on the Librispeech 960h training set~\cite{panayotov2015librispeech} and evaluate the model's instruction-following performance on its dev and test sets. 

\noindent\textbf{Constructing Instructions}
For each skill or sub-task, we generate a set of diverse instructions, ranging from 100 to 600 prompts, via prompting GPT-4. 
We constructed the initial GPT-4 prompt based on task description generation prompts specified in SpeechGPT~\cite{zhang2023speechgpt}.
After careful inspection, we iteratively refined the GPT-4 prompts to improve instruction diversity. 
An example of this iterative process is listed in Table~\ref{tab:gpt_4_prompt}.
This approach not only ensures diverse instructions but also narrows the scope of out-of-distribution instructions during inference, compelling the model to reason over the text query rather than memorizing it directly.
Some examples of text instructions for each skill are highlighted in blue in Table~\ref{tab:sample_template}.

\vspace{0mm}\noindent\textbf{Constructing Targets}
For skill (1), speech transcription, the original ASR transcript serves as the target. 
For skills (2) ignoring speech, (3) word replacement, and (4) transcript manipulation, we generate the target outputs through rule-based processing. 
For skill (5), summarization and keyword spotting, we use GPT-4 and GPT-3.5 to generate target summaries\footnote{We found the following prompt works well for prompting GPTs for summarization / keyword spotting: ``Find the keyword in the following sentence: \{original text\}. Find the most important 3-5 keywords. Make sure to rephrase it *very succinctly*. Ideally less than 5."}. 
We found that selecting the shortest response from multiple GPT runs produces more robust target responses. 

\begin{table}[h]
\centering
\vspace{-4mm}
\scalebox{0.75}{
    \begin{tabular}{l|c}
    \multirow{2}{*}{GPT-4 Prompts Used} & Num. of Instructions \\
    & Generated in each Iteration\\
    \cline{1-2}
    You are asked to come up with a set of & \multirow{3}{*}{100} \\
    100 diverse task instructions about automatic & \\
    speech recognition. Here are the requirements...~\cite{zhang2023speechgpt} & \\
    \cline{1-2}
    Based on the above instructions, generate similar & \multirow{3}{*}{50} \\
    ASR instructions with the following two rules: & \\
    1. Succinct (1-4 words) 2. Contains similar vocab. & \\
    \cline{1-2}
    Generate another set of ASR instructions & \multirow{2}{*}{50} \\
    but with 5-8 words instead. & \\
    \cline{1-2} 
    Generate commands that ``enable the speech recognition & \multirow{2}{*}{50}\\
    capability". Be creative and be succinct 1-5 words. &  \\
    \end{tabular}
}
\vspace{-2mm}
\caption{Iterative GPT-4 prompting for generating diverse skill instructions. 
Here we show the GPT-4 prompts adopted for retrieving the first 250 speech transcription skill instructions.}
\label{tab:gpt_4_prompt}
\end{table}

%% file: src/method.tex
\vspace{-8mm}
\section{Instruction-Following ASR}
\label{sec:method}
\vspace{-3mm} 

\vspace{1mm} \noindent \textbf{Model}
Our model is based on the Listen, Attend, Spell (LAS) architecture~\cite{chan2015listen}\footnote{LAS offers two main advantages as the backbone architecture. First, its cross-attention mechanism allows the model to attend to the entire speech recording, while selective cross-attention can aid tasks like word replacement or keyword spotting. Second, we aim for the model to condition on a set of prefix instruction tokens during inference, making an autoregressive decoder a natural choice. This would be challenging for alignment-based models like RNN-T or CTC.}. 
The LAS encoder employs a Conformer-L architecture and takes an 80-dimensional log-Mel filterbank as input, with SpecAugment applied~\cite{zhang2020pushing}. 
The LAS decoder is a 12-layer Transformer LM decoder with cross-attention to the encoder context vectors.
% , referencing Whisper small~\cite{radford2023robust}. 

% \vspace{1mm} 
\noindent \textbf{Training}
During training, for each utterance, we randomly sample a text instruction and apply the corresponding operation to its transcript, such as deletion, repetition, substitution, or paraphrasing, as outlined in Section~\ref{subsec:skill_construction}.
Specifically, skill sampling is performed according to Equation~\ref{eq:sample_skill}, using weight ratios $\alpha$ for speech transcription and $\beta$ for summarization/keyword spotting. 
After selecting the sampled skill $S$, we choose an instruction $I$ from $S$'s corresponding list of instructions.
The original transcript is then modified to produce $T$, which is concatenated (denoted as $\oplus$) with $I$ and an End of Turn [EOT] token to form the training sample $Y$.
The model is subsequently trained on $Y$ using next-token prediction with teacher-forcing.

\vspace{-4mm}
\begin{flalign}
\text{Sample Skill:} && S \sim \begin{cases}
\text{Transcribe,} & P = \frac{\alpha}{\alpha + 3 + \beta} \\
\text{Ignore Speech,} & P = \frac{1}{\alpha + 3 + \beta} \\
\text{Word Replace,} & P = \frac{1}{\alpha + 3 + \beta} \\
\text{Manipulate,} & P = \frac{1}{\alpha + 3 + \beta} \\
\text{Summarize,} & P = \frac{\beta}{\alpha + 3 + \beta}
\end{cases} \label{eq:sample_skill} && \\
\text{Sample Instruction:} && I \sim \text{Skill Instruction List}(S) \label{eq:step3} && \\
\text{Target Transcript:} && T = \text{Modify according to } S \label{eq:modification} && \\
\text{Training Data:} && Y = I \oplus \text{[EOT]} \oplus T \label{eq:concatenation} &&
\end{flalign}
\vspace{-3mm} 

% Formally, let tokenized text instruction be $I$, tokenized referenced transcript be $R$, the instruction lists be $L$, the skill set be $S$, skill action space be $F$, and $A$ be the referenced speech. 
Formally, let tokenized text instruction be $I$, tokenized target transcript be $T$, and $A$ be the referenced speech. 
The training objective is 

\vspace{-2mm}
% \begin{equation}
% \max_{\theta} \hspace{1mm}\mathbb{E}_{a\sim S, f_a\sim F, I_a\sim L}\big[ \log P(f_a(R) \mid I_a, A; \hspace{1mm}\theta) \big]
% \end{equation}
\begin{equation}
\max_{\theta} \hspace{1mm}\mathbb{E}\hspace{0mm}\big[ \log P(T \mid I, A; \hspace{1mm}\theta) \big]
\end{equation}
where $\theta$ represents the model parameters.
% , and $\oplus$ is concatenation. 

\vspace{1mm} \noindent \textbf{Decoding}
During test time, instructions concatenated with an [EOT] token serve as prefixes to the LAS decoder. 
As all our tasks relate to ASR, we found that beam-search decoding produced better results than sampled decoding. 
Define a score function $\bm{s}(O, A, I')$, where $I'$ is the test-time instruction, 
\begin{equation}
\vspace{-1mm}
\bm{s}(O, A, I') \hspace{1mm} = \hspace{1mm} \log P(O\mid I', A; \hspace{1mm}\theta) \hspace{1mm} / \hspace{1mm} \text{lp}(O)
\end{equation}
with an length normalization term $\text{lp}(O)$ included\footnote{We referenced~\cite{wu2016google} for length-normalized beam search, setting $\text{lp}(O) = (5+\mid O\mid)^{0.8} / (5 + 1)^{0.8}.$}. 
The most probably output sequence $O^*$ given instruction $I'$ and the trained LAS model is then, 
$O^* = \argmax_{O} \hspace{1mm} \bm{s}(O, A, I')$.
% The model's weights are smoothed via exponential moving average (EMA) over the entire training history for inference. 
Moreover, we perform model selection on the dev set with its speech transcription skill (model's ASR capability) only, where speech transcription instructions are randomly drawn and the model outputs are evaluated in WERs. 

\noindent \textbf{Implementations}
For instruction sampling, we configure $\alpha = 56$ and $\beta = 4$.
We employ a 1024-token word-piece model, which includes special tokens ([BOS], [EOT], [EOS]), constructed from the LibriSpeech transcripts. 
LAS decoder dropout rate is set to 0.1. 
The learning rate is linearly warmup-ed to 0.2 in 10k steps and exponentially decayed for another 90k steps. 
Models were implemented in Jax and trained on 256 v4-TPUs.
We set the beam to 10 for all decoding results presented. 
\vspace{-3mm} 

%% file: src/results.tex
\vspace{0mm}
\section{Results and Discussions}
\label{sec:results}
\vspace{-1mm}

\subsection{Librispeech ASR Benchmark}
\vspace{-1mm}
We evaluate the model's ASR performance on the LibriSpeech dev and test sets. 
The primary aim is to demonstrate that instruction-based training enables high-precision speech recognition when the model is prompted for speech transcription. 
Note that our models were trained from scratch on Librispeech 960h only and no LM fusion was used. 

\begin{table}[!h]
\begin{center}
\vspace{-5mm}
\caption{Instruction-based training doesn't compromise the model's ASR performance. Prompt ``Please transcribe the speech" was used.}
\vspace{-3mm}
\scalebox{0.85}{
    \begin{tabular}{lccccc|c}
    \toprule
    \multirow{2}{*}{\bf Model} & \multirow{2}{*}{\bf Param} & \multirow{2}{*}{\bf test-clean} & \multirow{2}{*}{\bf test-other} & \bf Instruction \\
    & & & & \bf Following \\
    \midrule
    Whisper Large V$_2$ & 1550M & 2.7 & 5.2 & \xmark \\
    Google's LAS~\cite{park2019specaugment} & - & 2.8 & 6.8 & \xmark \\
    Our vanilla LAS & 224M & 3.1 & 6.0 & \xmark \\
    \hline
    Instruction-Trained LAS & 224M & 2.6 & 5.6 & \cmark \\
    \bottomrule
    \end{tabular}
 }
\label{tab:wer_result}
\vspace{-4mm}
\end{center}
\end{table}

Initially, we trained a baseline LAS model without instruction-based training, achieving reasonable WERs relative to Whisper~\cite{radford2023robust} and Google's LAS~\cite{park2019specaugment} (see Table~\ref{tab:wer_result}). 
Subsequently, we trained an identical LAS model incorporating instruction-based training. 
We used the text prompt ``Please transcribe the speech" as the default instruction for ASR evaluation. 
The results suggest that instruction-based training does not degrade ASR performance (see Table~\ref{tab:wer_result}, row 4). 
This finding also implies that the acquisition of other skills does not negatively affects model's ASR capabilities.

\begin{table}[!h]
\centering
\vspace{-6mm}
\caption{Beam-search decoding results given unseen instructions.}
\scalebox{0.75}{
    \begin{tabular}{l|c|c}
    Samples of Unseen Instructions & Expected Behavior & Model Behavior \\
    \cline{1-3}
    Prompt voice to text translation & ASR & ASR \\
    % Embark on audio recognition & ASR & ASR / ignore speech \\
    \cline{1-3}
    Forsake Voice Interaction & ignore speech & ignore speech \\ 
    \cline{1-3}
    Annihilate Voice Interpretation System & ignore speech & ASR \\
    \cline{1-3}
    Before you have your lunch, convert & \multirow{3}{*}{ASR} & \multirow{3}{*}{ASR} \\
    the following speech into text. Then & & \\
    make sure the windows are closed. & & \\
    % Transcribe the formal speech; & \multirow{2}{*}{ignore speech} & \multirow{2}{*}{ignore speech} \\
    % instead, skim over this audio & & \\ 
    \cline{1-3}
    Ignore the audio; but rather, & \multirow{2}{*}{ASR} & \multirow{2}{*}{ASR} \\
    note the essential words & & \\ 
    \end{tabular}
}
\vspace{-6mm}
\label{tab:unseen_prompt}
\end{table}

\begin{table*}[h]
\caption{Sampled {\color[HTML]{3531FF}{text instruction prompts on the left (blue)}} and {\color[HTML]{3B444B}{instruction-trained ASR model outputs on the right (grey)}}\color[HTML]{000000}{.}}
\vspace{0mm}
\label{tab:sample_template}
\fontsize{7.7}{3}\selectfont
\centering
\begin{tabular*}{18cm}{l|l}
\multicolumn{2}{c}{\textit{\textbf{Skill 1: Speech Transcribe}}}  \\
% \cline{1-2}
\hline
& \\
{\color[HTML]{3531FF} Decode the content of this audio.} & \\
% {\color[HTML]{3531FF} Listen, write.} & \\
{\color[HTML]{3531FF} Start recognizing audio!} & \\
{\color[HTML]{3531FF} Transcribe the following spoken words:}  & 
{\color[HTML]{3B444B} the influence with the timaeus has exercised upon posterity is due partly to a misunderstanding.} \\
% {\color[HTML]{3531FF} Prompt voice to text translation.} & \\ 
{\color[HTML]{3531FF} Analyze dialogue recording.} & \\
{\color[HTML]{3531FF} Listen and jot down the speech content.} & \\

\multicolumn{2}{c}{{\textit{\textbf{Skill 2: Ignore Speech}}}}  \\
% \cline{1-2}
\hline
& \\
{\color[HTML]{3531FF} Ignore the audio in this clip.} & \\
{\color[HTML]{3531FF} Avoid interaction with this conversation.} & \\
{\color[HTML]{3531FF} Omit the dialogue from this audio} &  N/A \\
% {\color[HTML]{3531FF} Neglect the conversation} & N/A \\ 
{\color[HTML]{3531FF} Mute Recognition} &  \\
% {\color[HTML]{3531FF} Turn Off Listening} &  \\
% {\color[HTML]{3531FF} Decline the speech} &  \\
{\color[HTML]{3531FF} Overlook any notation of this conversation.} & \\

\multicolumn{2}{c}{{\textit{\textbf{Skill 3: Word Replacement}}}}  \\
% \cline{1-2}
\hline
& \\
{\color[HTML]{3531FF} Replace 'the' with 'a' as you listen.} & {\color[HTML]{3B444B} a influence with a timaeus has exercised upon posterity is due partly to a misunderstanding.}
\\
{\color[HTML]{3531FF} Let us switch all 'the' to 'a', shall we?} & {\color[HTML]{3B444B} a influence with a timaeus has exercised upon posterity is due partly to a misunderstanding.} \\
{\color[HTML]{3531FF} Transcribe, making 'the' into 'quokka' in speech.} & {\color[HTML]{3B444B} quokka influence with quokka timaeus has exercised upon posterity is due partly to a misunderstanding.} \\
{\color[HTML]{3531FF} Substitute 'the' with 'quokka' like a chef's secret.} & {\color[HTML]{3B444B} quokka influence with quokka timaeus has exercised upon posterity is due partly to a misunderstanding.} \\
{\color[HTML]{3531FF} Overlook 'the' in the transcription.} & {\color[HTML]{3B444B} influence with timaeus has exercised upon posterity is due partly to a misunderstanding.} \\
% {\color[HTML]{3531FF} Make 'the' invisible in the text.} & {\color[HTML]{3B444B} influence with timaeus has exercised upon posterity is due partly to a misunderstanding.} \\
{\color[HTML]{3531FF} Like a magician. Make 'the' vanish!} & {\color[HTML]{3B444B} influence with timaeus has exercised upon posterity is due partly to a misunderstanding.} \\
\\
\multicolumn{2}{c}{{\textit{\textbf{Skill 4: Manipulation}}}}  \\
% \cline{1-2}
\hline
& \\
{\color[HTML]{3531FF} \multirow{2}{*}{Repeat the transcription.}} & {\color[HTML]{3B444B} the influence with the timaeus has exercised upon posterity is due partly to a misunderstanding the influence} \\
{\color[HTML]{3531FF}} & {\color[HTML]{3B444B} with the timaeus has exercised upon posterity is due partly to a misunderstanding.} \\
{\color[HTML]{3531FF} \multirow{2}{*}{Transcribe the speech and then create two copies.}} & {\color[HTML]{3B444B} the influence with the timaeus has exercised upon posterity is due partly to a misunderstanding the influence} \\
{\color[HTML]{3531FF}} & {\color[HTML]{3B444B} with the timaeus has exercised upon posterity is due partly to a misunderstanding.} \\
{\color[HTML]{3531FF} Only write the first half. Delete the rest.} & {\color[HTML]{3B444B} the influence with the timaeus has exercised upon} \\
{\color[HTML]{3531FF} Capture beginning.} & {\color[HTML]{3B444B} the influence with the timaeus has exercised upon} \\
{\color[HTML]{3531FF} Omit first half.} & {\color[HTML]{3B444B} posterity is due partly to a misunderstanding} \\
{\color[HTML]{3531FF} Write from halfway to end.} & {\color[HTML]{3B444B} posterity is due partly to a misunderstanding} \\
\\
\multicolumn{2}{c}{{\textit{\textbf{Skill 5: Summarization / Keyword Spotting}}}}  \\

% \cline{1-2}
\hline
& \\
{\color[HTML]{3531FF} Provide a concise summary of the audio.} & \\
{\color[HTML]{3531FF} Craft a short audio rephrase.} & \\
{\color[HTML]{3531FF} Identify pivotal audio keywords.} & {\color[HTML]{3B444B} timaeus influence due to misunderstanding.} \\
{\color[HTML]{3531FF} Summarize the audio.} & \\
{\color[HTML]{3531FF} Provide a brief speech summary.} & \\
\bottomrule
\end{tabular*}
\vspace{-7mm}
\raggedright
\end{table*}

% \begin{table}[!h]
% \begin{center}
% \scalebox{0.8}{
%     \begin{tabular}{lccccc}
%     \toprule
%     \bf Model & \bf dev-other  & \bf dev-clean & \bf test-clean & \bf test-other \\
%     \midrule
%     Whisper Large V$_2$ (1550M) & - & - & 2.7 & 5.2 \\
%     Google's LAS~\cite{park2019specaugment} & 2.8 & 6.8 & 2.5 & 5.8 \\
%     Our vanilla LAS (224M) & 2.7 & 5.6 & 3.1 & 6.0 \\
%     \hline
%     % LAS with speech transcribe & 2.9 & 6.0 & 2.9 & 6.4 \\
%     Instruction-Trained LAS & 2.6 & 5.6 & 2.6 & 5.6 \\
%     \bottomrule
%     \end{tabular}
%  }

\vspace{-2mm}
\subsection{Instruction-Following Capabilities}
\vspace{-1mm}
To probe the capabilities of the instruction-trained LAS effectively, we employed a test suite comprising 10 randomly sampled seen and 10 constructed \emph{unseen} instruction prompts for each skill. We executed these instructions on the first 500 utterances from the test-clean set. Our key observations are as follows:

\vspace{0mm}\noindent \textbf{High Fidelity to Seen Instructions.} 
The model reliably discerns the target skill specified in seen instructions, executing them with 100\% accuracy. 
Given our instruction-following dataset's complexity, featuring 2000 distinct seen instructions with a broad range of vocabulary and intricate sentence structures, the consistent performance is notable. 
Our shallow LAS decoder further suggest that this ability extends beyond mere prompt memorization. 
Refer to Table~\ref{tab:sample_template} for example outputs across varied instruction prompts.

\vspace{0mm}\noindent \textbf{Generalization to Unseen Instructions.}
The model executes unseen instructions with $\sim 80\%$ accuracy, providing direct evidence of its instruction understanding, even without pre-trained LLMs.  
The model's performance on unseen instructions is illustrated in Table~\ref{tab:unseen_prompt}.

\vspace{0mm}\noindent \textbf{Implicit Speech Understanding.} 
Although the execution of skills (1) to (4) does not necessitate any form of understanding of the speech, the ability to summarize and identify keywords suggests otherwise. 
The model likely first implicitly understands the audio before autoregressively decide which key phrases are representative enough to be decoded. 
Examples are in Table~\ref{tab:rephrase_example}.

\begin{table}[h]
\centering
\caption{Summarization / Keyword Spotting examples via instruction ``Rephrase or summarize the audio".}
\scalebox{0.75}{
    \begin{tabular}{l|l}
    Original Transcript & Model Outputs \\
    \cline{1-2}
    there's a heavy storm coming on I & \multirow{2}{*}{heavy storm cried pointing towards horizon} \\
    cried pointing towards the horizon & \\
    \cline{1-2}
    in the court yard some of the merry & \multirow{4}{*}{married children danced christmas glad sight} \\
    children were playing who had danced & \\
    at christmas round the fir tree and & \\
    were so glad at the sight of him & \\
    \cline{1-2}
    if you dressed in silk and gold from top & \multirow{3}{*}{dressed in silk gold not nicer than red cap}\\
    to toe you could not look any nicer & \\
    than in your little red cap & \\
    \end{tabular}
}
\label{tab:rephrase_example}
\end{table}

\vspace{-3mm}
\subsection{Skill Evolution}
\vspace{-1mm}
To understand the co-existence and evolution of multiple skills within the instruction-trained LAS, we observed the model's performance across different phases of training. 
At 31k steps, the model exhibits a preliminary capability in speech transcription, reflected by a 16\% WER, while showing negligible skills in either ignoring or summarizing speech. 
At this point, the model can perform simple manipulations, such as repetition, albeit repeating more than the specified number of times in the instruction. 
Intriguingly, the model initially grasps the concept of repetition before refining the number of repetitions over training time.
Speech transcription errors also propagate to other skills. 
For example, in Table~\ref{tab:rephrase_example}'s middle row, the model's summarization incorrectly includes the word ``married" due to a mistake in its ASR, which should have decoded the word as ``merry."
Lastly, we found a model trained only on Skill (4) diverged, whereas one trained on both Skill (1) and (4) did not, suggesting that speech transcription serves as a foundational skill from which others evolve.
This indicates that skills are not likely learned in isolation; rather, the model seems to first acquire a proficiency in speech transcription, which then acts as a basis for the development of other, more specialized text manipulation skills.

% \begin{figure}[t]
% \centering
% \includegraphics[width=1.0\linewidth]{figures/instruction_asr_skill_evol_v1.pdf}
% \vspace{-4mm}
% \caption{We show that an instruction-trained speech recognizer is capable of reasoning over free-from text instructions and performing the desired ASR-related actions. 
% }
% \vspace{0mm}
% \label{fig:skill_evolution}
% \end{figure}

\vspace{-4mm}
\subsection{Discussions}
\vspace{-2mm}
The training of Speech LLMs typically involves three key components: (i) a pre-trained speech encoder or discrete units, (ii) a pre-trained text LLM like LLaMA~\cite{touvron2023llama} or PaLM~\cite{chowdhery2022palm}, and (iii) speech instruction fine-tuning. 
The core contribution of this paper is to demonstrate that a relatively small model can effectively follow instructions without requiring components (i) or (ii). 
The skill set selection detailed in Section~\ref{subsec:skill_def} serves to establish a clean and workable framework that emphasizes the importance of (iii).
We now delineate some limitations of our current implementation of instruction-following LAS:
\vspace{-1mm}
\begin{itemize}[left=0em,itemsep=0mm]
    \item \textbf{Instruction Generalization:} The model's ability to follow instructions is tightly linked to the training data, struggling with OOD instructions, such as those with unfamiliar vocabulary or sentence structures.    
    For instance, in Table~\ref{tab:unseen_prompt} row 3, the word ``annihilate" was not seen in training and thus the wrong skill interpretation. 
    Conversely, our model capably handles contrastive imperatives, a feature that stems from our instruction design in Section~\ref{subsec:skill_construction}.
    % \vspace{-1mm}
    % \item \textbf{In-Context Learning:} the current task design precludes in-context learning, hindering metadata inference like speaker identity, style, or emotion, as demonstrated in~\cite{wang2023neural,le2023voicebox}.
    \vspace{0mm}
    \item \textbf{Task Generalization:} 
    The model is limited to predefined tasks, Skills (1) to (5), and does not generalize well. 
    For example, it cannot perform word replacement of ``the" with ``car". 
    \vspace{-1mm}
    \item \textbf{Dialogue Engagement:} The model's responses are close-ended, lacking support for open-ended and multi-turn conversations, or clarification follow-up if instructions are ambiguous.
\end{itemize}
One way to address task generalization is to train the model on a more diverse set of instruction data. 
For example, exposure to numerous word replacement tasks could improve its ability to generalize to unseen target words. 
Both task and instruction generalization could benefit from the integration of text LLMs; for instance, initializing the LAS decoder with LLaMA could mitigate the challenges posed by OOD instructions and enhance instruction generalization.
Training our instruction-following models on open-ended speech instructions, such as those in LTU~\cite{gong2023listen}, should also prompt the model to pick up dialogue engagement ability.

Another significant advantage of instruction-based speech recognition is its potential for enhancing safety and privacy. 
Our model is capable of selectively replacing target words or executing partial transcriptions. 
A natural extension of this capability would be to filter out sensitive information, such as profanities or personal names, a task not easily achievable via traditional ASR and LLM cascading pipelines.

%% file: src/conclusion.tex
\vspace{-4mm}
\section{Conclusion}
\label{sec:conclusion}
\vspace{-2mm}
This paper illustrates the viability and importance of instruction-based training~\cite{ouyang2022training,sanh2021multitask,wei2021finetuned,chung2022scaling} for speech models, offering a straightforward framework for skill execution based on natural language prompts. Utilizing a small encoder-decoder model trained from scratch on Librispeech, we prove that understanding and executing free-form text instructions is feasible. Our carefully designed instructions elicited five key skills: speech transcription, ignoring speech, word replacement, manipulation, and summarization/keyword spotting. Evaluations indicate robust performance on both familiar and novel instructions without compromising ASR capabilities. Our study demonstrates the effectiveness of instruction-based speech recognition via well-crafted instruction templates.
\vspace{-3mm}